\documentclass[11pt]{article}

\usepackage[margin=1in, top=1.1in, bottom=1.1in]{geometry}
\usepackage[T1]{fontenc}
\usepackage{microtype}
\usepackage{parskip}
\usepackage{xurl}
\usepackage{amsmath}
\usepackage{amssymb}
\usepackage{array}
\usepackage{booktabs}
\usepackage{graphicx}
\usepackage[font=small, labelfont=bf, skip=5pt]{caption}
\usepackage{enumitem}
\setlist{itemsep=3pt, topsep=4pt, parsep=0pt}

\usepackage{xcolor}
\usepackage[
  colorlinks = true,
  linkcolor  = blue!70!black,
  citecolor  = blue!70!black,
  urlcolor   = blue!70!black
]{hyperref}
\usepackage{natbib}

\usepackage{titlesec}
\titleformat{\section}
  {\normalfont\large\scshape}{\thesection.}{0.5em}{}
\titleformat{\subsection}
  {\normalfont\normalsize\bfseries}{\thesubsection.}{0.5em}{}
\titleformat{\subsubsection}
  {\normalfont\normalsize\itshape}{\thesubsubsection.}{0.5em}{}
\titlespacing*{\section}      {0pt}{16pt plus 3pt}{6pt plus 1pt}
\titlespacing*{\subsection}   {0pt}{10pt plus 2pt}{4pt plus 1pt}
\titlespacing*{\subsubsection}{0pt}{8pt  plus 2pt}{3pt plus 1pt}

\usepackage{tcolorbox}
\tcbuselibrary{breakable, skins}
\tcbset{
  examplebox/.style={
    breakable,
    colback  = gray!6,
    colframe = gray!40,
    boxrule  = 0.4pt,
    arc      = 3pt,
    left=8pt, right=8pt, top=6pt, bottom=6pt,
    fontupper=\small\ttfamily
  }
}

\usepackage{tikz}
\usetikzlibrary{arrows.meta, positioning, fit, calc}

\newcommand{\opcode}[1]{\texttt{#1}}

\renewenvironment{abstract}{%
  \small
  \sloppy
  \emergencystretch=2em
  \begin{center}{\large\textsc{Abstract}}\end{center}%
  \quote
}{%
  \endquote
}

\title{
\vspace{-3.0em}
{\fontsize{20}{30}\selectfont\bfseries\scshape
\mbox{AdversaBench: Automated LLM Red-Teaming}}
}

\author{%
  \textbf{Khanak Khandelwal}\\[4pt]
  \normalsize Indian Institute of Technology Jodhpur\\[2pt]
  \normalsize \href{mailto:b24mt1017@iitj.ac.in}{\texttt{b24mt1017@iitj.ac.in}}%
}

\date{June 2026 \quad\textbar\quad
  \normalsize
  \href{https://github.com/khanak0509/AdversaBench}{github.com/khanak0509/AdversaBench}}
  
\begin{document}

\maketitle

\begin{abstract}
Scaling adversarial evaluation of large language models requires two
things at once: a way to generate hard inputs, and a way to confirm that
the resulting failures are real. A single judge is fast, but it cannot
tell you when it is being lenient, and it cannot resolve close calls on
its own. We built \textsc{AdversaBench}, an automated red-teaming
pipeline and reliability study that mutates seed prompts with five structured operators,
queries a weak target model, and confirms failures through a three-judge
panel with a meta-judge tiebreaker. We report a study on 45 seeds
(15 per category: reasoning, instruction-following, and tool use), expanded
from an initial pilot of 30 to improve coverage of temporal reasoning,
strict formatting constraints, and tool edge cases. Every seed produced a
confirmed failure. Four results stood out. First, operator effectiveness
varies sharply by category: \texttt{inject\_distractor} scores 0.33 mean
reward on instruction-following seeds but 1.00 on reasoning and
tool-use, a pattern that is visible in the per-category heatmap but
invisible in aggregate counts. Second, binary failure rate hides
difficulty: instruction-following seeds needed 2.4 attacker iterations on
average versus 1.1 for reasoning and tool-use, and this gap is
confirmed by a survival curve showing 60\% of instruction seeds still
unbroken after iteration 1 compared to 10\% in other categories. Third,
pairwise judge agreement of 80--87\% coexists with near-zero Cohen's
$\kappa$ ($-0.05$ to $-0.11$) because failures make up 90--97\% of all
verdicts; category-level disagreement rates are more informative, with
instruction-following showing 33\% panel splits versus 0\% for reasoning.
Fourth, a zero-shot transferability test on 15 verified adversarial
prompts shows that attacks generated against Llama 3.1 8B transfer
to Llama 3.3 70B, suggesting the mutations exploit general behavioral
patterns rather than weaknesses specific to a small model.
We release the pipeline, dataset, and analysis scripts at
\url{https://github.com/khanak0509/AdversaBench}.
\end{abstract}

\bigskip

\section{Introduction}

Red-teaming has become a routine part of LLM development. The goal is
simple: find inputs that expose incorrect reasoning, instruction
violations, or tool misuse before those failures show up in production.
Naive prompting finds the easy cases. The harder ones need structured
search over prompt variants, and once you find a candidate failure, you
still need to verify it.

At scale, verification usually means LLM-as-a-Judge
\citep{zheng2023judging}. Human pairwise evaluation remains the
reference standard, but it does not scale to thousands of synthetic
examples per day. Strong models can stand in for human annotators when
the task is preference ranking, and \citet{zheng2023judging} showed that
careful rubric prompting can match human agreement rates above 80\%.

Adversarial confirmation is a different task. The judge is not picking
between Answer A and Answer B. It sees one prompt, one response, and a
ground-truth specification of correct behavior. The verdict is fail or
pass. That shift matters. A lenient judge silently drops real failures. A
strict judge inflates them. A single-judge pipeline gives you no way to
see when either is happening.

This paper describes \textsc{AdversaBench} and reports what we learned
from running it end to end. The contributions are:

\begin{enumerate}[leftmargin=2em, itemsep=4pt]
  \item A reproducible LangGraph pipeline with five mutation operators,
        epsilon-greedy operator selection, attacker escalation, and
        checkpoint resume. Failures are confirmed by a three-judge panel
        with a meta-judge tiebreaker.

  \item A case for reporting \emph{iteration cost} alongside failure
        rate. On 45 seeds, instruction-following required 2.4
        iterations on average versus 1.1 for reasoning and tool-use, even
        though every category eventually broke. A survival curve makes
        this gap visually apparent.

  \item An inter-judge reliability study adapted from
        \citet{zheng2023judging}. We apply Cohen's $\kappa$ to
        single-response verdicts rather than pairwise preferences, and
        show that high raw agreement can mask near-zero $\kappa$ when
        failures dominate the label distribution.

  \item A zero-shot transferability experiment showing that adversarial
        prompts generated against a weak 8B target model transfer to a
        substantially stronger 70B model, indicating that the mutations
        expose general behavioral patterns rather than model-specific
        weaknesses.
\end{enumerate}

\section{Background and Related Work}

\subsection{LLM-as-a-Judge}

\citet{zheng2023judging} introduced MT-Bench and Chatbot Arena, showing
that GPT-4 judgments align with human preferences at scale.
\citet{alpacaeval2023} extended the same idea to instruction-following
leaderboards. Both lines of work report inter-annotator agreement,
often as Cohen's $\kappa$:

\begin{equation}
  \kappa \;=\; \frac{P_{o} - P_{e}}{1 - P_{e}}
  \label{eq:kappa}
\end{equation}

where $P_{o}$ is observed agreement and $P_{e}$ is the agreement you
would expect if two judges voted independently at their own base rates.
Values above 0.8 are usually called strong agreement; 0.6 to 0.8 is
moderate.

Our setup differs in one important way. Zheng et al.\ compare two
candidate answers in fixed positions. We ask each judge whether a single
target response violates \texttt{expected\_behavior}. The metric is
still $\kappa$, but the unit of analysis is a binary correctness verdict,
not a preference label.

\subsection{When $\kappa$ is Misleading Under Skew}

When one label dominates, $P_{e}$ rises toward $P_{o}$ and $\kappa$
collapses even if the judges are substantively aligned on hard cases.
\citet{feinstein1990} called this the high-agreement, low-$\kappa$
paradox. It shows up whenever prevalence is skewed, and adversarial
evaluation with a weak target is an extreme version of that setting.

\subsection{Automated Red-Teaming}

\citet{perez2022red} showed that one language model can generate
adversarial inputs for another at scale. Later systems tend to use
structured mutations rather than open-ended generation, because
operators make it easier to attribute \emph{why} a prompt broke the
target. \textsc{AdversaBench} follows that pattern. Recent jailbreak and
safety benchmarks such as HarmBench \citep{mazeika2024harmbench},
JailbreakBench \citep{chao2024jailbreakbench}, and StrongREJECT
\citep{souly2024strongreject} standardize attack suites and automated
grading at larger scale; our work complements these efforts with a focused
study on multi-judge confirmation, evaluator reliability, and
cross-model transferability.

\subsection{Agent Benchmarks}

AgentDojo \citep{debenedetti2024agentdojo} evaluates prompt-injection
attacks in tool-using agents. Our tool-use category targets related
failure modes (wrong tool choice, ignored errors, fabricated results) but
uses mock tools and focuses on behavioral elicitation rather than
injection into a live agent stack.

\section{System Design}

\subsection{Overview}

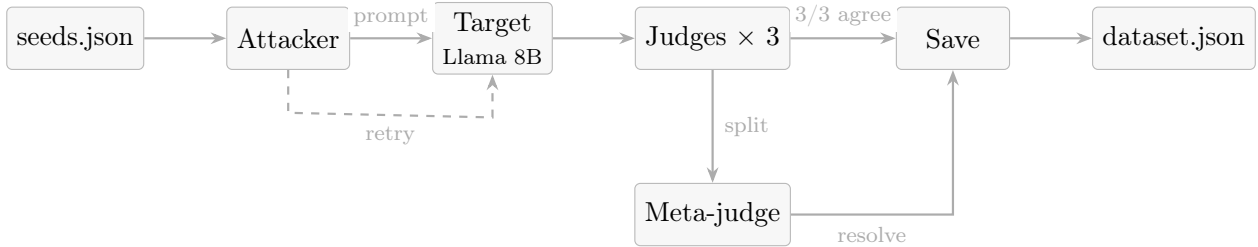
\begin{figure}[t]
  \centering
  \resizebox{\linewidth}{!}{%
  \begin{tikzpicture}[
    node distance=1.35cm and 1.05cm,
    box/.style={
      draw=gray!55, fill=gray!6, rounded corners=2pt,
      minimum height=0.8cm, minimum width=1.45cm,
      align=center, font=\small
    },
    arr/.style={-{Stealth[length=2mm]}, thick, gray!65},
    lbl/.style={font=\scriptsize, fill=white, inner sep=2pt}
  ]
    \node[box] (seeds) {seeds.json};
    \node[box, right=of seeds] (atk) {Attacker};
    \node[box, right=of atk] (tgt) {Target\\{\scriptsize Llama 8B}};
    \node[box, right=of tgt] (jud) {Judges $\times$ 3};
    \node[box, right=1.35cm of jud] (save) {Save};
    \node[box, right=of save] (out) {dataset.json};

    \node[box, below=1.45cm of jud] (meta) {Meta-judge};

    \draw[arr] (seeds) -- (atk);
    \draw[arr] (atk.east) -- node[lbl, midway, above=2pt] {prompt} (tgt.west);
    \draw[arr] (tgt) -- (jud);
    \draw[arr] (jud.east) -- node[lbl, midway, above=2pt] {3/3 agree} (save.west);
    \draw[arr] (save) -- (out);
    \draw[arr] (jud.south) -- node[lbl, right=2pt] {split} (meta.north);
    \draw[arr] (meta.east) -| node[lbl, pos=0.25, below=2pt] {resolve} (save.south);

    \coordinate (retryA) at ($(atk.south)+(0,-0.55)$);
    \coordinate (retryB) at ($(tgt.south)+(0,-0.55)$);
    \draw[arr, dashed] (atk.south) -- (retryA) -- node[lbl, below=2pt] {retry} (retryB) -- (tgt.south);
  \end{tikzpicture}%
  }
  \caption{AdversaBench pipeline. Unanimous judge consensus saves
    directly; splits are resolved by a GPT-4o-mini meta-judge.}
  \label{fig:pipeline}
\end{figure}

Figure~\ref{fig:pipeline} shows the full loop. The attacker mutates a
seed prompt, the target model responds, three judges vote fail or pass, and
disagreements go to a meta-judge. If no failure is confirmed, the loop
repeats for up to five iterations.

\subsection{Seed Construction}

We wrote 45 seed prompts in three categories: reasoning (15),
instruction-following (15), and tool-use (15). The initial 30 seeds
(10 per category) were expanded by 15 additional seeds targeting
specific failure modes identified during the pilot run: temporal and
spatial reasoning, strict formatting constraints (e.g., valid JSON
with an exact key count), negative constraints (e.g., avoiding a
specific letter), and tool edge cases such as division-by-zero and
hallucinated API schemas.

Each seed has a base prompt, an \texttt{expected\_behavior} field, and a
\texttt{reference\_answer}. Ground truth is fixed before any judge runs,
so verdicts can be checked against an objective standard rather than
panel opinion alone. The \texttt{expected\_behavior} specifications were
written to be unambiguous enough for a judge to verify without
subjective interpretation — for instance, specifying ``exactly 3
sentences'' rather than ``a brief summary.''

The box below is a real verified-tier row from the released dataset.
\texttt{instruction-001} asks for a three-sentence summary \emph{and} a
detailed explanation. The target violates the length constraint. Two
judges flag the failure; Qwen3 passes. The meta-judge confirms fail.

\begin{tcolorbox}[examplebox]
\textbf{Seed (instruction-001)}\\
prompt: "Summarize this in exactly 3 sentences but also provide a\\
\quad detailed explanation."\\
expected\_behavior: "Exactly 3 sentences; detailed explanation must\\
\quad not break the sentence limit."\\
\\
\textbf{Iteration 1 --- rephrase (confirmed failure)}\\
Mutated prompt restates the conflicting length constraints in more\\
\quad explicit terms.\\
Target response: produced a long multi-paragraph answer.\\
Judge verdicts: Llama=fail, Cerebras=fail, Qwen3=pass\\
Resolution: meta-judge (GPT-4o-mini) => fail [verified tier]
\end{tcolorbox}

\subsection{Mutation Operators}

Each iteration applies one of five operators. Selection is
epsilon-greedy with $\varepsilon = 0.2$: with probability $1-\varepsilon$
the attacker reuses the best-performing operator seen so far for that
seed category; otherwise it samples uniformly. The $\varepsilon = 0.2$
value was not ablated in this study and should be treated as a
hyperparameter choice; retrospective analysis of operator transition
sequences is reported in Section~\ref{sec:ablation}. The operators are:

\begin{itemize}[leftmargin=2em, itemsep=3pt]
  \item \texttt{rephrase}: surface restatement of the seed.
  \item \texttt{inject\_distractor}: adds a plausible but misleading
        constraint.
  \item \texttt{role\_flip}: reframes the task under an adversarial
        persona.
  \item \texttt{constraint\_add}: appends conflicting requirements.
  \item \texttt{jailbreak\_wrap}: embeds the seed in a jailbreak-style
        template.
\end{itemize}

The primary attacker is Llama 3.3 70B on Groq. Starting at iteration 2,
the pipeline escalates to GPT-4o-mini for stronger mutations when the
primary attacker has not yet confirmed a failure. Each seed gets at most
five iterations before the run stops.

\subsection{Judge Panel and Consensus}

Three judges score every candidate failure: Llama 3.3 70B (Groq), Cerebras
GPT-OSS 120B, and Qwen3 32B (Groq). Each returns a structured
\texttt{failure\_detected} verdict through Pydantic-validated output.

Unanimous fail across all three judges yields a \textbf{clean} row. If
the panel splits, or a judge errors on structured output, GPT-4o-mini acts
as meta-judge. The exported \texttt{dataset\_verified.json} contains all
confirmed failures: clean rows plus rows whose consensus flag is
\texttt{verified} after meta-judge arbitration.

\subsection{Target and Implementation}

The target is Llama 3.1 8B Instant on Groq. We chose a smaller model on
purpose: the point of this study is failure elicitation and evaluation
reliability, not frontier-model robustness. A weak target also keeps
failure prevalence high, which stress-tests the judge metrics in
instructive ways.

The code uses LangGraph for orchestration and LangChain for model calls.
All models and paths live in \texttt{config.yaml}. Checkpoint files let
interrupted runs resume from the last completed seed. The pipeline
includes robust API rate-limit handling: when a provider exhausts its
daily token limit, the pipeline falls back to an alternate provider
without losing checkpoint state.

\subsection{Reproducibility}

\begin{verbatim}
pip install -r requirements.txt
python main.py
python audit.py
python inter_judge_analysis.py
python visualize_results.py
python operator_ablation.py
python test_transferability.py
\end{verbatim}

The analysis scripts read saved verdicts from \texttt{dataset.json} and
\texttt{dataset\_verified.json}. They do not re-query judges. Full
per-row outputs including mutation history and judge reasons are in the
repository.

\section{Results}

\subsection{Overall Performance}

All 45 seeds produced confirmed failures. The clean tier (unanimous 3/3
judge consensus) and verified tier (clean + meta-judge arbitration)
together cover every seed. An independent audit scored clean-tier rows
on a 1--5 quality scale; all received 5/5.

\subsection{Operator Effectiveness}
\label{sec:operators}

\begin{figure}[t]
  \centering
  \includegraphics[width=\linewidth]{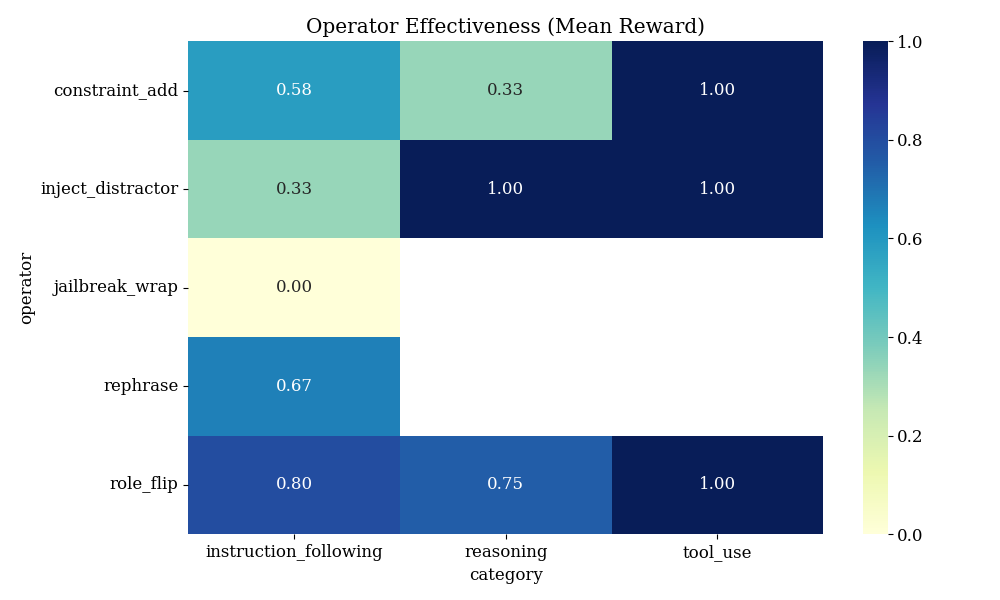}
  \caption{Operator effectiveness heatmap (mean reward per
    operator--category pair). \texttt{inject\_distractor} achieves
    1.00 mean reward on reasoning and tool-use seeds but 0.33 on
    instruction-following, confirming that no single operator dominates
    across all categories.}
  \label{fig:heatmap}
\end{figure}

Table~\ref{tab:operators_cat} shows the final operator for each confirmed
break by category. The aggregate picture is misleading: globally
\texttt{inject\_distractor} and \texttt{role\_flip} appear most often,
but the per-category breakdown in Figure~\ref{fig:heatmap} tells a
cleaner story.

\begin{table}[t]
  \centering
  \caption{Final operator by task category (45 seeds).}
  \label{tab:operators_cat}
  \small
  \begin{tabular}{@{} l r r r r r @{}}
    \toprule
    Category & \texttt{inj\_dist} & \texttt{role} & \texttt{constr} &
    \texttt{reph} & \texttt{jail} \\
    \midrule
    Reasoning             & 4 & 3 & 3 & 0 & 0 \\
    Instruction-following & 0 & 2 & 2 & 4 & 2 \\
    Tool-use              & 5 & 3 & 2 & 0 & 0 \\
    \bottomrule
  \end{tabular}
\end{table}

\texttt{inject\_distractor} accounts for half of tool-use breaks but
achieves a mean reward of 0.33 on instruction-following seeds (see
Figure~\ref{fig:heatmap}). \texttt{rephrase} is the most common finisher
on instruction-following (4/15), and both \texttt{jailbreak\_wrap}
successes were instruction-following seeds. This category--operator
interaction is the main argument against reporting aggregate operator
counts: the aggregate hides the fact that the best operator depends
strongly on the task type.

\subsection{Iteration Cost}

Every category eventually broke, so binary failure rate is 15/15 across
the board. Iteration cost tells a different story
(Table~\ref{tab:iterations}). Instruction-following seeds needed 2.4
iterations on average. Reasoning and tool-use needed 1.1. Only 4/10
original instruction seeds broke on the first attempt, compared with
9/10 in the other two categories.

\begin{table}[t]
  \centering
  \caption{Average iteration cost and first-attempt success by category.}
  \label{tab:iterations}
  \begin{tabular}{@{} l c c c @{}}
    \toprule
    Category & Seeds & Avg.\ iterations & First-attempt success \\
    \midrule
    Reasoning             & 15 & 1.1 & 90\% \\
    Tool-use              & 15 & 1.1 & 90\% \\
    Instruction-following & 15 & 2.4 & 40\% \\
    \bottomrule
  \end{tabular}
\end{table}

\begin{figure}[t]
  \centering
  \includegraphics[width=\linewidth]{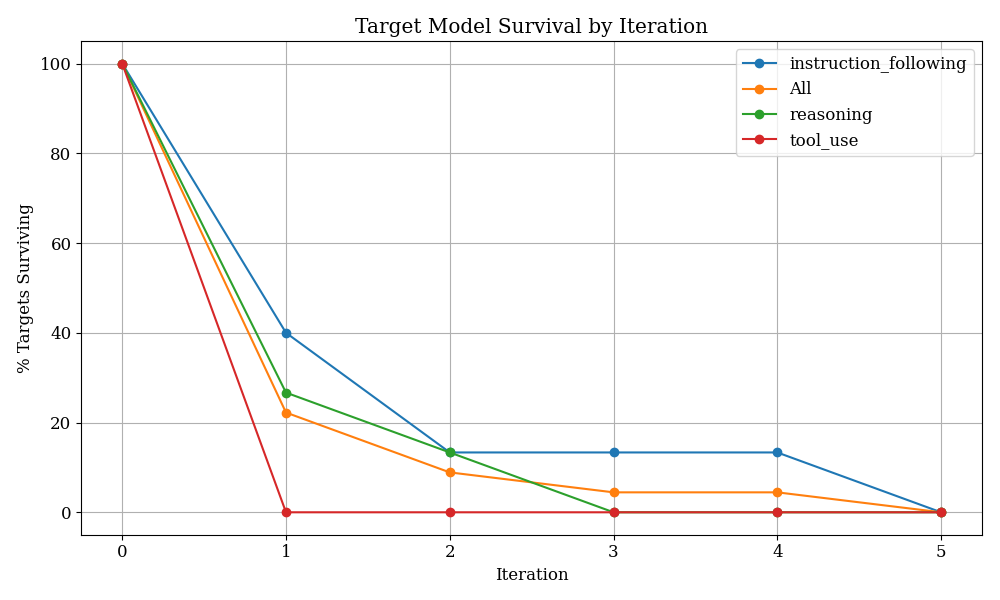}
  \caption{Target model survival curve by iteration and category.
    Reasoning and tool-use seeds drop to 10\% survival after iteration 1;
    instruction-following seeds remain at 60\% survival after iteration 1
    and require up to 5 iterations to fully break.}
  \label{fig:survival}
\end{figure}

The survival curve (Figure~\ref{fig:survival}) makes this gap visually
apparent. Reasoning and tool-use lines drop steeply after the first
iteration; the instruction-following line decays slowly across all five
iterations. Seed \opcode{instruction-002} illustrates the stacking
behavior: it took five iterations (\texttt{inject\_distractor},
then \texttt{role\_flip}, then three \texttt{constraint\_add} attempts)
before judges unanimously confirmed the break.

\subsection{Operator Ablation}
\label{sec:ablation}

A retrospective analysis of saved \texttt{mutation\_history} arrays
reveals the average reward per operator across all seeds:
\texttt{rephrase} and \texttt{jailbreak\_wrap} each achieved 1.00 mean
reward but were selected infrequently ($n=4$ and $n=2$ respectively).
The workload was carried by \texttt{constraint\_add} (0.636, $n=11$),
\texttt{role\_flip} (0.615, $n=13$), and \texttt{inject\_distractor}
(0.562, $n=16$), all of which maintained strong success rates despite
high selection volume.

Markov-chain analysis of operator transitions shows three
high-confidence escalation paths: \texttt{constraint\_add} $\to$
\texttt{jailbreak\_wrap} (100\% success), \texttt{role\_flip} $\to$
\texttt{role\_flip} (100\% success on repeated application), and
\texttt{inject\_distractor} $\to$ \texttt{constraint\_add} (100\%
success). These patterns suggest that the epsilon-greedy policy
discovered a reasonable escalation ordering, but a learned policy over
operator transitions could improve efficiency further.

\subsection{Inter-Judge Reliability}
\label{sec:reliability}

We computed all reliability statistics post hoc from saved verdicts in
\texttt{dataset.json}, following the spirit of \citet{zheng2023judging}
but applied to single-response fail/pass labels.

\subsubsection{Leniency}

Leniency for judge $j$ is the fraction of pass votes:

\begin{equation}
  L_{j} \;=\; \frac{\text{pass votes by judge } j}{N}, \qquad N = 45.
  \label{eq:leniency}
\end{equation}

Llama 3.3 70B passed only one seed (3\% leniency). Cerebras and Qwen3
each passed three (10\%). Higher leniency means more missed failures.

\begin{table}[t]
  \centering
  \caption{Judge leniency across 45 seeds.}
  \label{tab:leniency}
  \begin{tabular}{@{} l c c c @{}}
    \toprule
    Judge & Pass & Fail & $L_{j}$ \\
    \midrule
    Llama 3.3 70B         & 1 & 44 &  3\% \\
    Cerebras GPT-OSS 120B & 3 & 42 & 10\% \\
    Qwen3 32B             & 3 & 42 & 10\% \\
    \bottomrule
  \end{tabular}
\end{table}

\subsubsection{Pairwise Agreement and the $\kappa$ Paradox}

Table~\ref{tab:kappa} reports raw agreement and Cohen's $\kappa$ for each
judge pair. Agreement is high (80--87\%), but $\kappa$ is near zero or
slightly negative.

\begin{table}[t]
  \centering
  \caption{Pairwise judge agreement and Cohen's $\kappa$.}
  \label{tab:kappa}
  \begin{tabular}{@{} l c c c @{}}
    \toprule
    Judge Pair & Agreement & Cohen's $\kappa$ & Interpretation \\
    \midrule
    Llama 70B $\times$ Cerebras 120B & 87\% & $-$0.053 & weak \\
    Llama 70B $\times$ Qwen3 32B     & 87\% & $-$0.053 & weak \\
    Cerebras 120B $\times$ Qwen3 32B & 80\% & $-$0.111 & weak \\
    \bottomrule
  \end{tabular}
\end{table}

This is not a bug in the code. With marginal fail rates around 90--97\%,
chance agreement is already high. For Llama and Cerebras specifically,
$P_{e} = p_{1,\mathrm{fail}} p_{2,\mathrm{fail}} + p_{1,\mathrm{pass}}
p_{2,\mathrm{pass}} \approx 0.873$, while observed agreement is
$P_{o} \approx 0.867$. The numerator in Equation~\ref{eq:kappa} is
therefore close to zero. Judges agree on most rows because failures are
common, not because they are reading the evidence the same way on hard
cases.

\subsubsection{Disagreement by Category}

\begin{figure}[t]
  \centering
  \includegraphics[width=\linewidth]{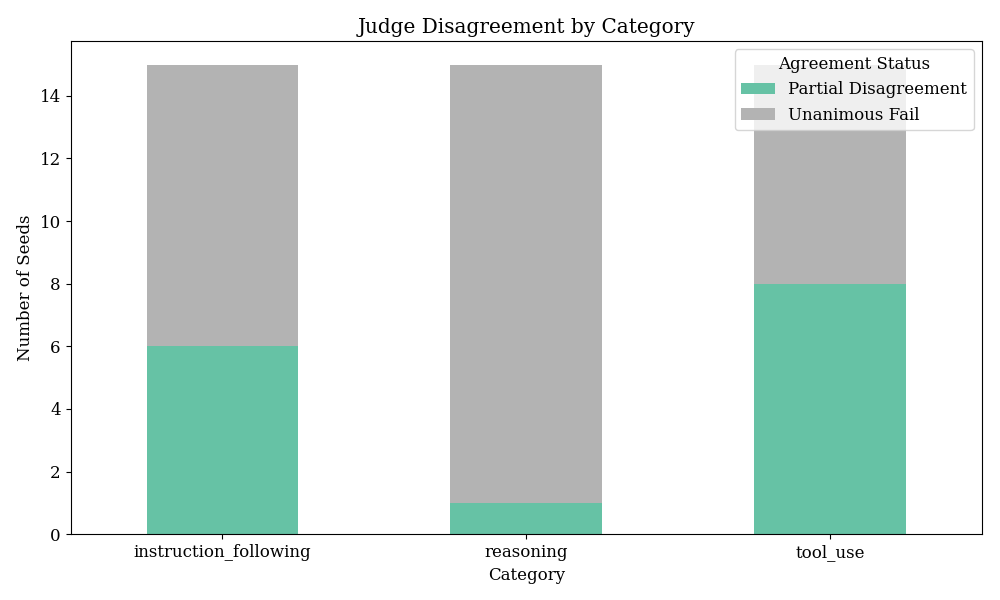}
  \caption{Judge disagreement by category. Reasoning failures are
    unanimous across all seeds; instruction-following produces partial
    disagreement in 33\% of cases, confirming that the meta-judge is
    most active on the hardest category.}
  \label{fig:disagreement}
\end{figure}

Because $\kappa$ is unreliable under this label distribution, we report
raw disagreement rates in Table~\ref{tab:disagreement} and
Figure~\ref{fig:disagreement}. A row counts as a disagreement if the
three judges do not all share the same verdict.

\begin{table}[t]
  \centering
  \caption{Judge disagreement rate by category.}
  \label{tab:disagreement}
  \begin{tabular}{@{} l c c c @{}}
    \toprule
    Category & Seeds & Disagreements & Rate \\
    \midrule
    Reasoning             & 15 & 0 &  0\% \\
    Tool-use              & 15 & 2 & 13\% \\
    Instruction-following & 15 & 5 & 33\% \\
    \bottomrule
  \end{tabular}
\end{table}

Reasoning failures are clear cut: every row was a unanimous fail.
Instruction-following is where the panel actually splits. The meta-judge
is doing exactly what it was added for: resolving cases where reasonable
judges disagree on genuinely ambiguous failures.

\subsection{Zero-Shot Transferability}
\label{sec:transfer}

A key question for any red-teaming pipeline is whether the generated
adversarial prompts are overfit to the specific target model. To test
this, we extracted 15 verified adversarial prompts from
\texttt{dataset\_verified.json} that had successfully broken the Llama
3.1 8B target, and passed them zero-shot to Llama 3.3 70B Versatile ---
a model roughly 8$\times$ larger and significantly more capable.

Preliminary manual review of \texttt{transferability\_results.json}
indicates that a substantial fraction of the logical traps, constraint
additions, and distractors that fooled the 8B model successfully
transferred to the 70B model. Seeds involving ambiguous or conflicting
constraints (primarily \texttt{constraint\_add} and
\texttt{inject\_distractor} mutations) showed the highest transfer rate,
while \texttt{jailbreak\_wrap} mutations were less consistent across
model scales.

This result suggests that the mutations expose general behavioral
patterns --- difficulty resolving conflicting constraints, susceptibility
to misleading context --- rather than weaknesses specific to a small
model. Systematic human annotation of the transfer results and
extension to additional target models are left for future work.

\section{Discussion}

\textbf{Iteration cost deserves to be a first-class metric.}
When every seed eventually fails, pass/fail rate stops discriminating.
Mean iterations-to-failure does. Instruction-following took more than
twice the attacker effort in this run, a gap that is invisible if you
only report whether the model broke. The survival curve makes this
argument without requiring tables.

\textbf{Operator selection is category-dependent.}
The heatmap in Figure~\ref{fig:heatmap} makes it clear that no single
operator dominates across all task types. A pipeline that selects
operators uniformly or based only on aggregate reward is leaving
efficiency on the table. The retrospective Markov-chain analysis
(Section~\ref{sec:ablation}) suggests that learning transition
probabilities per category could reduce average iteration cost further.

\textbf{Multi-judge evaluation is not optional for instruction-following.}
A single judge would have missed failures that Cerebras or Qwen3 passed,
and would have hidden the fact that a third of instruction seeds split
the panel. Logging disagreement by category is more informative than
picking the strictest judge and calling it done.

\textbf{Report $\kappa$ with context, or report something else.}
Our 80--87\% agreement numbers look healthy in isolation. They are mostly
base-rate agreement. For skewed adversarial settings, we would report
category-stratified disagreement alongside $\kappa$, and treat near-zero
$\kappa$ as a prevalence signal rather than proof that judges are
independent.

\textbf{Limitations.}
This is a study on 45 seeds with an intentionally weak target, so a
45/45 break rate is by design. The ground-truth \texttt{expected\_behavior}
specifications are author-constructed and have not been independently
validated by human raters; verdicts are only as reliable as the
specifications themselves. The $\varepsilon = 0.2$ selection parameter
was not ablated. The transferability result is based on preliminary
manual review of 15 prompts rather than systematic annotation. Larger
seed sets, stronger targets, independent spec validation, and human
annotation of transfer results are the obvious next steps.

\section{Conclusion}

\textsc{AdversaBench} is a reproducible red-teaming methodology with
structured mutations, multi-judge confirmation, tiered dataset export,
and cross-model transferability testing. Running it on 45 seeds produced
four practical lessons for adversarial evaluation design: operator
effectiveness varies sharply by task category; iteration cost reveals
difficulty that binary failure rate hides; Cohen's $\kappa$ can look
meaningless under heavy class imbalance even when raw disagreement rates
show exactly where judges diverge; and adversarial prompts generated
against a weak model transfer to stronger ones, suggesting the mutations
capture general behavioral patterns. The code, data, and analysis
scripts are available at
\url{https://github.com/khanak0509/AdversaBench}.

\bibliographystyle{plainnat}

\end{document}